\documentclass[envcountsame]{mmnp}
\usepackage[tbtags]{amsmath}
\usepackage{amssymb}
\usepackage{graphicx}
\usepackage[latin1]{inputenc}

\usepackage{enumitem}
\usepackage[justification=centering]{caption}

\numberwithin{equation}{section}

\begin{document}

\title{Fuzzy Adaptive Resonance Theory, Diffusion Maps and their applications to Clustering and Biclustering}

\author{S.~B.~Damelin \inst{1} \sep Y.~Gu \inst{2} \sep D.~C.~Wunsch II \inst{3} \sep R.~Xu \inst{4} \thanks{\email {damelin@umich.edu}.}}

\vspace{0.5cm}

\institute{\inst{1} Mathematical Reviews \\
			 The American Mathematical Society \\
			 Ann Arbor, MI 48103 USA \\
              \inst{2} Department of Mathematics \\
	      		 University of Michigan -- Ann Arbor \\
			 Ann Arbor, MI 48109 USA \\
	      \inst{3} Applied Computational Intelligence Laboratory \\
	      		 Department of Electrical and Computer Engineering \\
			 University of Missouri -- Rolla \\
			 Rolla, MO 65409-0249 USA \\
	      \inst{4} GE Global Research \\
	                  Niskayuna, NY 12309 USA \\
			 \vspace{0.5cm}
	      Dedicated to our friend and colleague Prof.~Alexander Gorban.
}


\abstract{In this paper, we describe an algorithm FARDiff (Fuzzy Adaptive Resonance Diffusion) which combines Diffusion Maps and Fuzzy Adaptive Resonance Theory to do clustering on high dimensional data. We describe some applications of this method and some problems for future research.}





\titlerunning{ART, Diffusion Maps and App.~to Clustering and Biclustering}

\maketitle


\setcounter{equation}{0}
\section{Introduction}
In this paper, we describe an algorithm FARDiff (Fuzzy Adaptive Resonance Diffusion) which combines Diffusion Maps and Fuzzy Adaptive Resonance Theory to do clustering on high dimensional data. This algorithm was introduced in the papers \cite{b:9, b:10, b:18, b:19, b:21}. We describe some applications of this method and some problems for future research.

The dimensionality reduction part of FARDiff is achieved via a nonlinear diffusion map, which interprets the eigenfunctions of Markov matrices as a system of coordinates on the dataset in order to obtain an efficient representation of certain geometric descriptions of the data. Our algorithm is sensitive to the connectivity of the data points. Clustering was achieved using Fuzzy Adaptive Resonance Theory. 

The structure of this paper is as follows. Section 2 will describe the dimension reduction, section 3 will describe the clustering method, section 4 will discuss some applications, and section 5 will discuss work in progress.

\setcounter{equation}{0}
\section{Diffusion Maps}
In this section, we describe the nonlinear dimension reduction spectral method part of FARDiff. We use nonlinear diffusion maps. See \cite{b:6, b:15} and references sited there in.

Let $m$ be a large enough fixed non-negative integer, and let $\mathbf{X} = \{\mathbf{x}_{i}, \, i = 1,\dots,N\} \subset \mathbb{R}^{m}$ be a data set consisting of $N \geq 1$ distinct points. A finite graph with $N$ nodes corresponding to $N$ data points can be constructed on $\mathbf{X}$ as follows. Every two nodes in the graph are connected by an edge weighted through a Gaussian kernel, $w: \mathbb{R}^{m} \times \mathbb{R}^{m} \to \mathbb{R}$, defined by 
\[
	w(\mathbf{z},\mathbf{z'}) = \exp \left( - \frac{\big\| \mathbf{z} - \mathbf{z'} \big\|^{2}}{\sigma^{2}} \right), \qquad \mathbf{z},\,\mathbf{z'} \in \mathbb{R}^{m},
\]
where $\|\cdot\|$ is the Euclidean norm in $\mathbb{R}^{m}$ and $\sigma$ is a fixed nonzero real number to be chosen in a moment. $w$ here satisfies symmetric property $w(\mathbf{z},\mathbf{z'}) = w(\mathbf{z'},\mathbf{z})$ and positivity preserving property $w(\mathbf{z},\mathbf{z'}) \geq 0$. The first property is useful for performing spectral analysis, and the second property allows it to be interpreted as a scaled probability.

We choose $\sigma$ to be a parameter depending on the data set $\mathbf{X}$, and we shall write for each $1\leq i,j\leq N$,
\begin{equation}
	\label{eq:2.1}
	w(\mathbf{x}_{i},\mathbf{x}_{j}) = \exp \left( - \frac{\big\| \mathbf{x}_{i} - \mathbf{x}_{j} \big\|^{2}}{\sigma^{2}} \right), \qquad \mathbf{x}_{i},\,\mathbf{x}_{j} \in \mathbf{X}.
\end{equation}
$w$ reflects the degree of similarity between $\mathbf{x}_{i}$ and $\mathbf{x}_{j}$. The resulting symmetric semi-positive definite matrix $\mathbf{W} = \{w(\mathbf{x}_{i},\mathbf{x}_{j})\}_{N \times N}$ is called the affinity matrix. 

Let $d: \mathbf{X} \to \mathbb{R}$ be defined by
\begin{equation}
	\label{eq:2.2}
	d(\mathbf{x}) = \sum_{\mathbf{x'} \in \mathbf{X}} w(\mathbf{x},\mathbf{x'}), \qquad \mathbf{x} \in \mathbf{X},
\end{equation}
be the degree of $\mathbf{x}$. For each $1\leq i,j\leq N$, a Markov or affinity matrix $\mathbf{P}_{N \times N}$ is then constructed by calculating each entry of $\mathbf{P}$ as
\begin{equation}
	\label{eq:2.3}
	p(\mathbf{x}_{i},\mathbf{x}_{j}) = \frac{w(\mathbf{x}_{i},\mathbf{x}_{j})}{d(\mathbf{x}_{i})}.
\end{equation}
From the definition of $w$, $p(\mathbf{x}_{i},\mathbf{x}_{j})$ can be interpreted as the transition probability from $\mathbf{x}_{i}$ to $\mathbf{x}_{j}$ in one time step. From the definition of the Gaussian kernel it can be seen that the transition probability will be high for similar elements. This idea can be further extended by considering $p^{t}(\mathbf{x}_{i},\mathbf{x}_{j})$ in the $t^{\text{th}}$ power $\mathbf{P}^{t}$ of $\mathbf{P}$ as the probability of transition from $\mathbf{x}_{i}$ to $\mathbf{x}_{j}$ in $t$ time steps \cite{b:6}, where $t$ is a fixed non-negative integer. Here, $p^{t}(\mathbf{x}_{i},\mathbf{x}_{j})$ sums all paths of length $t$ from point $\mathbf{x}_{i}$ to point $\mathbf{x}_{j}$. Hence, the parameter $t$ defines the granularity of the analysis. With the increase of the value of $t$, local geometric information of data is also integrated. The change in size of $t$ makes it possible to control the generation of more specific or broader clusters.

The diffusion distance is related to $\mathbf{P}$ and is given by
\begin{equation}
	\label{eq:2.4}
	D_{t}(\mathbf{x}_{i},\mathbf{x}_{j})^{2} = \sum_{\mathbf{x} \in \mathbf{X}} \Big| p^{t}(\mathbf{x}_{i},\mathbf{x}) - p^{t}(\mathbf{x}_{j},\mathbf{x}) \Big|^{2}.
\end{equation}
It can be seen that the more paths that connect two points in the graph, the higher probability of paths of length $t$ between two points is, and the smaller the diffusion distance is. Therefore, the diffusion distance can be used as a measurement of a specific geometric structure.

Next, we aim to find a mapping that reassembles the data in a lower dimensional space but preserves the diffusion distances between data points optimally. Consider a mapping
\begin{equation}
	\label{eq:2.5}
	\boldsymbol{\Psi}'_{t}(\mathbf{x}_{i}) = \Big(p^{t}(\mathbf{x}_{i}, \mathbf{x}_{1}), \dots, p^{t}(\mathbf{x}_{i}, \mathbf{x}_{N})\Big)^{T}, \qquad 1 \leq i \leq N.
\end{equation}
The Euclidean distance between two points $\mathbf{x}_{i}$ and $\mathbf{x}_{j}$ is
\[
	\big\| \boldsymbol{\Psi}'_{t}(\mathbf{x}_{i}) - \boldsymbol{\Psi}'_{t}(\mathbf{x}_{j}) \big\|^{2} = \sum_{\mathbf{x} \in \mathbf{X}} \Big| p^{t}(\mathbf{x}_{i},\mathbf{x}) - p^{t}(\mathbf{x}_{j},\mathbf{x}) \Big|^{2},
\]
which is also the diffusion distance. The mapping thus preserves the original data structure in the sense of diffusion distance.

Next, we need to perform dimensionality reduction. Because of the symmetry property of $w$, for each $t$, we may obtain a sequence of $N$ eigenvalues of $\mathbf{P}$, $1 = \lambda_{1} > \lambda_{2} > \dots > \lambda_{N} \geq 0$, with corresponding eigenvectors $\{\boldsymbol{\Phi}_{j}, j = 1,\dots,N\}$, satisfying,
\begin{equation}
	\label{eq:2.6}
	\mathbf{P}^{t}\boldsymbol{\Phi}_{j} = \lambda^{t}_{j}\boldsymbol{\Phi}_{j}.
\end{equation}
Suppose we wish to reduce our data from $\mathbb{R}^{m}$ to $\mathbb{R}^{L}$ where $1 \leq L < m$\footnote{Note that here we require that $N\geq L$. This is not a strong requirement given the reduced dimension $L$ is typically small and the number of points are always larger.}. To achieve it, we choose the first $L$ eigenvalues and their corresponding eigenvectors of $\mathbf{P}$, and rewrite the mapping \eqref{eq:2.5} as
\begin{equation}
	\label{eq:2.7}
	\boldsymbol{\Psi}_{t}(\mathbf{x}_{i}) = \Big( \lambda_{1}^{t}\boldsymbol{\Phi}_{1}(\mathbf{x}_{i}),\dots,  \lambda_{L}^{t}\boldsymbol{\Phi}_{L}(\mathbf{x}_{i}) \Big)^{T},
\end{equation}
where $\boldsymbol{\Phi}_{j}(\mathbf{x}_{i})$ is the $i^{\text{th}}$ element of the eigenvector $\boldsymbol{\Phi}_{j}$. We observe that the new data point $\mathbf{x}_{i}$'s dimension has been reduced to $L$. The new mapping reduces the dimensionality of the original data set by retaining the important geometric structure associated with dominant eigenvalues and eigenvectors. Correspondingly, the diffusion distance between a pair of points $\mathbf{x}_{i}$ and $\mathbf{x}_{j}$ is approximated with the Euclidean distance in $\mathbb{R}^{L}$, written as
\begin{equation}
	\label{eq:2.8}
	D_{t}(\mathbf{x}_{i},\mathbf{x}_{j})^{2} = \Big\| \boldsymbol{\Psi}_{t}(\mathbf{x}_{i}) - \boldsymbol{\Psi}_{t}(\mathbf{x}_{j})  \Big\|^{2}.
\end{equation}


The kernel width parameter $\sigma$ represents the rate at which the similarity between two points decays\footnote{Our choice of $\sigma$ is determined by a trade off between sparseness of the kernel matrix (small $\sigma$) with adequate characterization of true affinity of two points.}. One of the main reasons for using spectral clustering methods is that, with sparse kernel matrices, long range affinities are accommodated through the chaining of many local interactions as opposed to standard Euclidean distance methods - e.g.~correlation - that impute global influence into each pair-wise affinity metric, making long range interactions dominate local interactions.

We have used a Gaussian Kernel scaled with parameter $\sigma$. This kernel works well for the two applications we have studied given both  trade sparseness and affinity of the point set. FARDIFF however can be defined by way of a wide class of positive definite kernels, see \cite{b:6, b:7} where the choice of kernel is typically application dependent.  In addition to the choice of kernel, the trade off in sparsity can be handled by using a restricted isometry, see for example \cite{b:1}.

\setcounter{equation}{0}
\section{Fuzzy Adaptive Resonance Theory}
In this section we describe the second part of FARDiff which uses Fuzzy Adaptive Resonance Theory (FA) \cite{b:3, b:17, b:20} for clustering data points whose dimension has been reduced using the method of section 2. See \cite{b:9, b:10, b:18, b:19, b:21}.

\begin{figure}[t]
	\centerline{\includegraphics[scale=0.5]{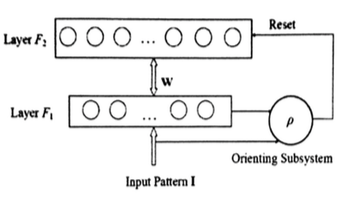}}
	\caption{Topological structure of Fuzzy ART. Layers $F_{1}$ and $F_{2}$ are connected via adaptive weights $\mathbf{W}$. The orienting subsystem is controlled by the vigilance parameter $\rho$.}
	\label{f:1}
\end{figure}

FA allows stable recognition of clusters in response to both binary and real-valued input patterns with either fast or slow learning. The basic FA architecture consists of two-layer nodes or neurons, the feature representation field $F_{1}$, and the category representation field $F_{2}$, as shown in Figure \ref{f:1}. The neurons in layer $F_{1}$ are activated by the input pattern, while the prototypes of the formed clusters are stored in layer $F_{2}$. The neurons in layer $F_{2}$ that are already being used as representations of input patterns are said to be committed. Correspondingly, the uncommitted neuron encodes no input patterns. The two layers are connected via adaptive weights $\mathbf{w}_{j}$, emanating from node $j$ in layer $F_{2}$. After an input pattern is presented, the neurons (including a certain number of committed neurons and one uncommitted neuron) in layer $F_{2}$ compete by calculating the category choice function
\begin{equation}
	\label{eq:1.9}
	T_{j} = T_{j}(\mathbf{x}, \mathbf{w}_{j}, \alpha) = \frac{\big| \mathbf{x} \wedge \mathbf{w}_{j} \big|}{\alpha + \big| \mathbf{w}_{j} \big|},
\end{equation}
where $\wedge$ is the fuzzy AND operator defined as follows. Given $\mathbf{y} = \{\mathbf{y}_{i}, \, i = 1,\dots,N\} \subset \mathbb{R}^{m}$ and $\mathbf{y'} = \{\mathbf{y'}_{i}, \, i = 1,\dots,N\} \subset \mathbb{R}^{m}$, define
\begin{equation}
	\label{eq:1.10}
	\big( \mathbf{y} \wedge \mathbf{y'} \big)_{i} = \min_{i} \big( \mathbf{y}_{i}, \mathbf{y}'_{i} \big),
\end{equation}
and $\alpha > 0$ is the choice parameter to break the tie when more than one prototype vector is a fuzzy subset of the input pattern, based on the winner-take-all rule,
\begin{equation}
	\label{eq:1.11}
	T_{J} = \max_{j} \{ T_{j} \}.
\end{equation}
The winning neuron $J$ then becomes activated, and an expectation is reflected in layer $F_{1}$ and compared with the input pattern. The orienting subsystem with the pre-specified vigilance parameter $\rho$ $(0 \leq \rho \leq 1)$ determines whether the expectation and the input pattern are closed matched. If the match meets the vigilance criterion,
\begin{equation}
	\label{eq:1.12}
	\rho \leq \frac{\big|  \mathbf{x} \wedge \mathbf{w}_{j} \big|}{\big| \mathbf{x} \big|},
\end{equation}
weight adaptation occurs, where learning starts and the weights are updated using the following learning rule,
\begin{equation}
	\label{eq:1.13}
	\mathbf{w}_{j}(\mathrm{new}) = \beta\big( \mathbf{x} \wedge \mathbf{w}_{j}(\mathrm{old}) \big) + \big(1-\beta\big)\mathbf{w}_{j}(\mathrm{old}),
\end{equation}
where $\beta \in [0,1]$ is the learning rate parameter. On the other hand, if the vigilance criterion is not met, a reset signal is sent back to layer $F_{2}$ to shut off the current winning neuron, which will remain disabled for the entire duration of the presentation of this input pattern, and a new competition is performed among the rest of the neurons. This new expectation is then projected into layer $F_{1}$, and this process repeats until the vigilance criterion is met. In the case that an uncommitted neuron is selected for coding, a new uncommitted neuron is created to represent a potential new cluster.

Some of ART's (Adaptive Resonance Theory) advantages are stability, biological plausibility, and responsiveness to the stability-plasticity dilemma. Further advantages are scalability, speed, configurability, and potential for parallelization. ART has been found to have good ability to interpret well, results on neural net learning and decision based networks with low complexity and good robustness \cite{b:16}.

\setcounter{equation}{0}
\section{Applications in Cancer Detection and Hyperspectral Clustering}
The algorithm which we discuss in this paper which we call FARDiff (Fuzzy Adaptive Resonance Diffusion) combines 1.~Diffusion Maps and 2.~Fuzzy Adaptive Resonance Theory to do clustering on high dimensional data. This algorithm was introduced in the papers \cite{b:9, b:10, b:18, b:19, b:21}.

In \cite{b:18, b:21}, we applied FARDiff to investigate cancer detection. Early detection of the site of the origin of a tumor is particularly important for cancer diagnosis and treatment. The employment of gene expression profiles for different cancer types or subtypes has already shown significant advantages over traditional cancer classification methods. We applied FARDiff to the small round blue-cell tumor (SRBCT) data set, which is published from the diagnostic research of small round blue-cell tumors in children, and compared results with other widely-used clustering algorithms. The experimental results demonstrate the effectiveness of our method in extracting useful information from the high-dimensional data sets.
 
In \cite{b:9, b:10, b:19}, we applied FARDiff to study hyperspectral imaging. The images are sampled at hundreds of frequencies, and the bands are regularly spaced, thus a continuous spectrum can be drawn for every pixel in the image. We reassembled our data into a ``hypercube,'' called the hyperspectral data. Due to the fact that each pixel of the image contains more than a hundred bands, the occurrence of large amounts of hyperspectral data brings important challenges to storage and processing. We used FARDiff to investigate clustering of high dimensional hyperspectral image data from core samples provided by AngloCold Ashanti, and compared to results obtained by AngloGold Ashanti's proprietary method. The experimental results demonstrate the potential of our method in addressing the complicated hyperspectral data and identifying the minerals in core samples.

\setcounter{equation}{0}
\section{Open Questions For Future Research}
In this section, we discuss some on going work. We are interested in extending the FARDiff algorithm to a biclustering framework and high dimensional data reduction. Biclustering is a technique which performs simultaneous clustering in many dimensions automatically integrating feature selection to clustering without any prior information. Two examples of good biclustering algorithms are BARTMAP and HBiFAM (Hierarchical Biclustering FA algorithm) \cite{b:16, b:22}.

It is well known that clustering has been used extensively in the analysis of high-throughput messenger RNA (mRNA) expression profiling with microarrays. This technique is restrictive, in part, due to the existence of many uncorrelated genes with respect to sample or condition clustering, or many unrelated samples or conditions with respect to gene clustering. Biclustering offers a solution to such problems by performing simultaneous clustering on both dimensions, or automatically integrating feature selection to clustering without any prior information, so that the relations of clusters of genes (generally, features) and clusters of samples or conditions (data objects) are established. Challenges which need to be addressed using this method include computational complexity and high dimensional data reduction. Current work focuses on developing a natural framework for an analog of FARDiff for biclustering and related computational complexity challenges, for example, traveling salesman problems. A second question relates to applications of FARDiff to more generalized clustering performance. A third research project investigates unified learning schemes and hardware implementation with ART and FARDiff in clustering and biclustering. A valuable new area of innovation will be the application of FARDiff to more generalized data structures such as trees and grammars. Continued progress on distributed representations is valuable because of increased data representation capability, both in terms of system capacity and template complexity. Several aspects of these questions are being investigated in ongoing work.



\begin{acknowledgement}
Damelin gratefully acknowledges support from the School of Computational and Applied Mathematics at Wits, the American Mathematical Society, the Center for High Performance Computing and the National Science Foundation. Xu and Wunsch gratefully acknowledge support from the Missouri University of Science \& Technology Intelligent Systems Center, and the M.~K.~Finley Missouri Endowment. Wunsch additionally acknowledges support from the National Science Foundation. Gu gratefully acknowledges support from the University of Michigan.
\end{acknowledgement}




\begin{thebibliography}{99}

\bibitem{b:1}
E.~J.~Cand\'{e}s and T.~Tao. Near-optimal signal recovery from random projections: universal encoding strategies. \emph{IEEE Trans.~Inform.~Theory}, (52)(2006), pp 5406-5425.

\bibitem{b:2} 
G.~Carpenter, S.~Grossberg, N.~Markuzon, J.~Reynolds, and D. Rosen, Fuzzy ARTMAP: A neural network architecture for incremental supervised learning of analog multidimensional maps, \emph{IEEE Transactions on Neural Networks}, (3)(1992), pp 698-713.

\bibitem{b:3} 
G.~Carpenter, S.~Grossberg, and D.~Rosen, Fuzzy ART: Fast Stable Learning and Categorization of Analog Patterns by an Adaptive Resonance, \emph{Neural Networks}, (4)(1991), pp 759-771.

\bibitem{b:4}
K.~Cawse, S.~B.~Damelin, R.~McIntyre, M.~Mitchley, L.~du Plessis and M.~Sears, An Investigation of data compression for Hyperspectral core image data, \emph{Proceedings of the Mathematics in Industry Study Group, South Africa}, 2008, pp 1-25.

\bibitem{b:5}
K.~Cawse, M.~Sears, A.~Robin, S.~B.~Damelin, K.~Wessels, F.~van den Bergh, R.~Mathieu, Using random matrix theory to determine the number of endmembers in a hyperspectral image, \emph{Proceedings of WHISPERS 2010, Reykjavik, Iceland}, pp 45-52.

\bibitem{b:6} 
R.~Coifman and S.~Lafon, Diffusion maps. \emph{Journal of Applied and Computational Harmonic Analysis}, April 2006, pp 5-30.

\bibitem{b:7}
S.~B.~Damelin, J.~Levesley, D.~L.~Ragozin and X.~Sun, Energies, Group Invariant Kernels and Numerical Integration on Compact Manifolds, \emph{Journal of Complexity}, (25)(2009), pp 152-162.

\bibitem{b:8}
S.~B.~Damelin and W.~Miller, Mathematics and Signal Processing, \emph{Cambridge Texts in Applied Mathematics}, no.~48, February 2012.

\bibitem{b:9} 
L.~du Plessis, R.~Xu, S.~B.~Damelin, M.~Sears and D.~C.~Wunsch, Reducing dimensionality of hyperspectral data with diffusion maps and clustering with K-means and fuzzy art, \emph{Int.~J.~Systems Control and Communications}, (3)(2011), pp 232-251.

\bibitem{b:10} 
L.~du Plessis, R.~Xu, S.~B.~Damelin, M.~Sears and D.~C.~Wunsch, Reducing dimensionality of hyperspectral data with diffusion maps and clustering with K-means and fuzzy art, \emph{Proceedings of IJCNN 2009}, pp 32-36.

\bibitem{b:11}
C.~Fefferman, S.~B.~Damelin and W.~Glover, \emph{BMO Theorems for $\varepsilon$ distorted diffeomorphisms on $\mathbb R^D$ and an application to comparing manifolds of speech and sound}, Involve 5-2(2012), pp 159-172.

\bibitem{b:12} 
S.~Madeira and A.~Oliveira, Biclustering algorithms for biological data analysis: A survey, \emph{IEEE Transactions on Computational Biology and Bioinformatics}, 1-1(2004), pp 24-45.

\bibitem{b:13}
M.~Mitchley, M.~Sears and S.~B.~Damelin, Target detection I Hyperpectral mineral data using wavelet analysis, \emph{Proceedings of the 2009 IEEE Geosciences and Remote Sensing Symposium, Cape Town}, pp 23-45.

\bibitem{b:14} 
S.~Mulder and D.~Wunsch II, Million city traveling salesman problem solution by divide and conquer clustering with adaptive resonance neural networks, \emph{Neural Networks}, (16)(2003), pp 827-832.

\bibitem{b:15}
J.~de la Porte, B.~M.~Herbst, W.~Hereman, S.~J.~van der Walt, An Introduction to Diffusion Maps, \emph{The 19th Symposium of the Pattern Recognition Association of South Africa}. 2008.

\bibitem{b:16}
D.~Wunsch II, ART Properties of Interest in Engineering Applications, \emph{Proceedings of International Conference of Neural Networks, Atlanta}, 2009.

\bibitem{b:17} 
R.~Xu and D.~Wunsch II. Clustering. \emph{IEEE/Wiley}, 2008.

\bibitem{b:18} 
R.~Xu, S.~B.~Damelin, and D.~C.~Wunsch II, Applications of diffusion maps in gene expression data-based cancer diagnosis analysis. In \emph{Proceedings of the 29th Annual International Conference of IEEE Engineering in Medicine and Biology Society, Lyon, France}, 2007, pp 4613-4616.

\bibitem{b:19} 
R.~Xu, L.~du Plessis, S.~B.~Damelin, M.~Sears, and D.~C.~Wunsch II, Analysis of Hyperspectral Data with Diffusion Maps and Fuzzy ART. In \emph{Proceedings of the 2009 international joint conference on Neural Networks}, pp 2302-2309.

\bibitem{b:20} 
R.~Xu and D.~Wunsch II, Survey of Clustering Algorithms, \emph{IEEE Transactions on Neural Networks}, 16-3(2005), pp 645-678.

\bibitem{b:21} 
R.~Xu, S.~B.~Damelin, B.~Nadler, and D.~C. Wunsch II, Clustering of High-Dimensional Gene Expression Data with Feature Filtering Methods and Diffusion Maps, \emph{Bio-Medical Engineering and Informatics}, (1)(2008), pp 245-249.

\bibitem{b:22}
R.~Xu and D.~C. Wunsch II, BARTMAP: A Viable Structure for Biclustering, \emph{Neural Networks}, (24)(7)(2011), pp 709-716.

\end{thebibliography}
\end{document}